%% file: ijcai18.tex
\documentclass{article}
\usepackage{ijcai18}

\usepackage{times}
\usepackage{xcolor}
\usepackage{soul}
\usepackage[utf8]{inputenc}
\usepackage[small]{caption}
\usepackage{natbib, url}
\usepackage{threeparttable}
\usepackage{tabularx}
\usepackage{booktabs,siunitx}
\usepackage{enumitem}

\usepackage{xspace}
\usepackage{subfig}
\usepackage{amssymb}
\usepackage{pifont}

\usepackage[colorlinks,linktocpage,draft]{hyperref}
\usepackage{url}
\usepackage{relsize}

\hypersetup{pdfpagemode=UseNone}  

\usepackage{graphicx}

\input{glyphtounicode}  
\usepackage{color} 
\definecolor{darkred}{rgb}{0.55, 0.0, 0.0}
\definecolor{orange}{RGB}{255,127,0}
\definecolor{brown}{RGB}{150,70,0}
\definecolor{green}{RGB}{127,255,127}
\definecolor{darkgreen}{RGB}{0,127,0}
\definecolor{blue}{RGB}{127,127,255}
\definecolor{lightblue}{RGB}{150,150,255}
\definecolor{darkblue}{RGB}{0,0,127}
\definecolor{red}{RGB}{255,90,90}
\definecolor{grey}{RGB}{127,127,127}
\definecolor{pink}{RGB}{255,180,180}

\title{A multidisciplinary task-based perspective for evaluating the impact of \\AI autonomy and generality on the future of work}

\author{
Enrique Fernández-Macías$^1$, 
Emilia Gómez$^{1,2}$, 
José Hernández-Orallo$^{3}$, \\
{\bf Bao Sheng Loe$^4$, 
Bertin Martens$^1$,
Fernando Mart\'inez-Plumed$^3$, 
Song\"ul Tolan$^1$}\thanks{Corresponding author. Authors listed in alphabetical order.} 
\\ 
$^1$ Joint Research Centre, European Commission, 
$^2$ Universitat Pompeu Fabra, \\
$^3$ Universitat Politècnica de València,
$^4$ University of Cambridge\\
\small{\{enrique.fernandez-macias,emilia.gomez-gutierrez,
bertin.martens,songul.tolan\}@ec.europa.eu}\\
\small{\{jorallo,fmartinez\}@dsic.upv.es},
\small{bsl28@cam.ac.uk}
}

\begin{document}

\maketitle

\begin{abstract}
  This paper presents a multidisciplinary task approach for assessing the impact of artificial intelligence on the future of work. We provide definitions of a task from two main perspectives: socio-economic and computational.
We propose to explore ways in which we can integrate or map these perspectives, and link them with the skills or capabilities required by them, for humans and AI systems. Finally, we argue that in order to understand the dynamics of tasks, we have to explore the relevance of autonomy and generality of AI systems for the automation or alteration of the workplace.
\end{abstract}

\section{Introduction}

Much attention is being paid on the impact of artificial intelligence (AI) in the future of work \citep{Kaplan2015,Frey2017,Arntz2016,Nedelkoska2018}. However, current research focuses on narrow views of the problem, i.e. they lack a comprehensive understanding about which and how tasks are (and will) be performed by state-of-the-art AI systems, 
how occupations are structured into tasks in different sectors, and 
what is the expected impact on the labour market. 

In many ways, the current impact of AI in the workplace can be seen as a new take on pushing the frontiers of scientific management \citep{taylor1914principles},  whose main objectives are to increase economic efficiency and labour productivity through the engineering process of automation and management. However, the importance of human attributes such as cognitive and personality involvement as determinants of performance at the workplace \citep[e.g.][]{cote2006emotional, barrick1991big, hunter1986cognitive} has been largely ignored in some recent studies \citep{Frey2017,Arntz2016,Nedelkoska2018}. Lesser attention is also given to the role of group dynamics in the workplace relating to the principles of communication, coordination and culture. Does the replacement of humans with AI truly increase job productivity for all workplaces? How are these interrelated principles measured in an environment where AI is becoming more dominant in the workplace? Each of these principles must be clearly defined within the realm of automation in order for it to be successful. Furthermore, the current view of AI in the future of work does not appear to address the relevance of learning capabilities for autonomously acquiring new skills and their ability to generally transfer their knowledge and experience to a range of tasks much more easily than other more narrow AI models.
\footnote{As the terms will be used throughout the paper, we give the following working definitions of autonomy and generality, without any intention of being normative beyond the scope of this paper. {\em Autonomy} is the capability of a system to determine and pursue its own subgoals and to modify its behaviour through optimisation, learning, self-programming or other kinds of development or adaptation. In other words, an autonomous system can pursue goals that were not programmed and in ways that were not programmed originally. {\em Generality} is the degree of coverage of a wide range of tasks (up to a level of resources). A specialised system can cover a few (possibly very complex) tasks, but a general system is expected to cover a wide range of (perhaps simpler) tasks.}

In this position paper we propose a multidisciplinary perspective to address this problem. We first define and distinguish differences between tasks, skills and occupations. 
We then discuss notions of ``task" from different perspectives: socio-economic and computational. 
We consider the misalignment between both conceptions and the pressures and distortions of each of them. We then consider whether skills or abilities can help bridge these different notions, and how they can be analysed ---and measured--- for humans and AI systems in a coherent way. We pay special attention on generality and autonomy which are key requirements for human labour in most work situations and how we can measure the extent to which AI systems are ready for those situations, or how fast we are progressing towards them.

As a result, we bring some insights into the way AI systems can replace or change the way humans perform certain tasks, 
which will guide us to an understanding of the impact of the advancements and integration of current AI systems into real-world job scenarios. 

\section{Tasks, skills and occupations}

The aim of this section is to clarify the distinction between tasks, skills and occupations in the workplace. Parts of this section rely on \citet{Fernandez2017}, where a more detailed description of the task, skills and occupations can be found. This section will also describe the impact of human abilities in the workplace.   

\subparagraph*{Tasks:}
According to the main proponent of the tasks approach in labour economics, tasks can be simply defined as units of work activity that produce output \citep{Autor2013}. The point of departure of this approach is a strictly technical view of production, seen as a mechanical process of transforming inputs into outputs. Work is an input in this process, and tasks are more or less discrete and distinct units of work. Depending on the complexity of the production process, it may require the combination of more or less different types of tasks, in the same way as it may require different types of raw materials. 

This poses two challenges to the definition of tasks: (1) discretisation and (2) level of granularity. The way we define a process that transforms an output from an input often involves the combination of different tasks. This makes it difficult to separate one task from the other. Nevertheless, a framework that enables the distribution of tasks between humans and machines requires a closed-form definition of a task. In addition, the level of granularity should be such that the process of transforming inputs to outputs should involve one closed task. However, human perception of the granularity of tasks is limited at the micro as well as macro level of processes that transform inputs to outputs.


\subparagraph*{Skills:} 
%
Skills are defined as the stock of human capabilities that allow human beings to perform tasks \citep{Autor2013}. It is a combination of  human abilities, experience and knowledge put together to perform an activity in a competent manner. Often, they are referred to as the specific psychomotor processes that are necessary to meet current requirements of a specific task and are manifested through human behaviours \citep{cheney1990knowledge}. Skills also facilitate the capability of the human being to select among a repertoire of possible actions that is deemed most appropriate in a particular situation.  

The classification of skills can be domain-specific (e.g skills necessary to survive at a workplace) or domain-general (e.g., soft skills vs hard skills). Regardless of how they are classified, task completion is skill dependent. While some tasks require  complex skills (e.g., managing a company), others require only very specialised skills (e.g., designing a car's engine).

There are certainly attributes of a person that can affect the way skills are learned and developed over time. For example, several studies have shown that performance of skill acquisition can be affected by personality, motivation and cognitive-intellectual determinants of individual differences \citep{ackerman1988determinants,kanfer1989motivation, bastian2005emotional}. These studies indicate that human beings acquire skill expertise at different levels based on various psychological attributes or abilities. We will discuss abilities next. 

Given a particular task, humans can either perform the task if one has already acquired the skills to do so, or they may require specialised training first to learn the skills in order to complete the task successfully. The level of which the individual learns the skills is dependent on several key factors. Often, the aptitude of the individual and his or her general cognitive ability determines the achievement level of the skills acquired. Previous research has shown that general cognitive ability is an important predictor of job performance \citep{schmidt2002role}, where the criteria of job performance are typically based on the extent to which the tasks assigned to the individual are executed successfully.

Another important indicator of human attribute to task performance is the personality traits and individuals' fit at work \citep{barrick1991big, kristof2005consequences}. An individual may have an aptitude for the task, but if the environment at the workplace is not compatible with the characteristics of the individual, then it is likely that he or she will struggle to perform well at the tasks.

It is thus, the responsibility of the human resource department to enhance human capital and ensure that candidates selected for the job are predicted to accomplish the tasks at high levels \citep{adkins1994judgments}. This is often achieved by measuring the job candidate's general cognitive ability, job-related skills, motivations and personality traits using various psychometric tests as a gauge of possible achievement levels \citep{morgeson2005importance}.



\subparagraph*{Occupations:}   
Theoretically occupations are defined as bundles of tasks that require a particular combination of skills, and correspond to positions within the social structure of productive organisations. However, the real-world application of the occupation definition is arbitrary and heavily influenced by regulations, historical reasons and social convention. Moreover, occupations under the same name (e.g., lawyer) may be just grouped by a domain or area of knowledge that the professionals in that area share, rather than by a coherent grouping of the tasks or skills involved.
Also, the tasks within an occupation may vary significantly in time, especially in the way they are executed, but the occupation is considered a more stable label.

%
%

\vspace{0.2cm}
\noindent
As we will discuss later on, the level of analysis is completely different according to tasks or occupations. For instance, analysing the probability that some occupations can be automated is very different from --but of course related to-- analysing what tasks have a highest probability of being automated.

%
%

\section{Different perspectives of what a task is}

While skills, abilities and occupations can help us understand and organise work, it is ultimately analysed in terms of what has to be done, i.e. the tasks. However, the notion of tasks varies in different perspectives. In this section, we present tasks from two perspectives : a socio-economic perspective as considered in the social science literature on the work process \citep[see e.g.][]{Braverman1974} and a computational perspective in terms of how the task is seen from the point of view of computerisation.
%
%

\subsection{The socio-economic perspective}

In the social sciences, tasks, skills, abilities and occupations have been categorised according to many different criteria. On the basis of a recent literature review on tasks and skills, \citet{Fernandez2017} propose a classification of tasks according to their role in the work process across two different dimensions: the contents of the task (what is the objective of the task) and the tools and methods used for carrying out the task (how the task is performed). This is illustrated in Table \ref{tab:classtask}, already presented in \citet{Fernandez2017}. 

\begin{table}[h!]
\centering
\small
\caption{A classification of tasks according to their contents and methods}
\input{casstasktab}

\label{tab:classtask}
\begin{tablenotes}[flushleft]
\item {\footnotesize Source: \citet{Fernandez2017}}
\end{tablenotes}
\end{table}

The task content dimension is sub-divided according to the object upon which the task is performed: physical tasks (that operate on things), intellectual tasks (that operate on information) and social tasks (that operate on people)\footnote{Sometimes, the term `manual' is used instead of `physical', `digital' instead of `intellectual`, and `service' instead of social, but all are partially cognitive (e.g., they cannot be replaced by an animal or a tool without some kind of complex information processing).}.

The second dimension of the task classification is subdivided into methods (work organisation), including autonomy, teamwork and routine, and tools (technology), including the use of basic tools, basic ICT or even programming. 


This second dimension of the task typology reflects the social organisation and technology used in production, and is therefore more historically and institutionally contingent (for the production of the same type of goods or services, different societies or organisations can use different technologies or forms of work organisation). However, the categories of tasks identified in this second dimension are also very relevant for automation and artificial intelligence research. For instance, a significant part of the literature considers that the degree of routine involved in a task is the crucial determinant of its automatability, even if this degree of routine is the result of a particular form of work organisation and not an attribute of the task content as such.
Thus, a particular (re)organisation of a task can be a precondition for its automation. For instance, the automation of manufacturing industries in the 20th Century was preceded by Taylorism and Scientific Management, which radically transformed the organization of work in factories by an extreme division of labour, standardisation of processes and reduction in the autonomy of workers.

Moreover, there are different transmission channels that are affected when a human task is replaced by a purely mechanical or digital algorithm-driven machine. Machines can substitute and/or complement human workers. In both cases they may trigger labour productivity increases for (remaining) workers and/or reduce output prices. The latter may have in-sector demand effects that may more than compensate the decline in labour per unit of output. There may also be cross-sector spill-over effects as higher wages fuel overall demand for products. Finally, employment and wage effects may results in an overall re-allocation of workers across sectors and the elasticity of labour supply will in turn determine the size of the implications for wages and employment. 

A large part of the economic literature on technology and automation revolves around the traditional capital/labour (K/L) substitution model with factor augmenting technological progress. That model leaves the possibility for increasing employment with capital deepening as long as the substitution elasticity is lower than 1.  A major breakthrough came with \citet{Acemoglu2016}, who propose a task-based model, which replaces factor augmentation with direct substitution between human and (automated) machine-executed tasks. This approach can model how increases in worker productivity due to automation will not lead to a proportional expansion of the demand for labour. 

The history of technological progress since the industrial revolution suggests a positive impact of technological progress on employment and wages. Despite massive technological progress and substitution of human labour by machines across nearly all sectors, unemployment has not increased while incomes have substantially increased. Empirical evidence for the traditional (K/L)-type model yields positive employment effects towards a skill-biased technical change. However, evidence for the task-based approach shows rather negative effects of automation on employment. These negative results are not surprising, given the emphasis of the task-based model on the displacement effect and the focus of the corresponding empirical literature on the manufacturing sectors which is prone to routinisation and automation. Thus, the empirical literature on the impact of automation/robotisation on the labour market is scarce and not decisive. 

\subsection{The computational perspective}

Computer science originally conceived tasks as transformations from inputs to outputs. 
This paradigm is still captured by sophisticated tasks using AI, such as object recognition, audio/visual synthesis or even translation. However, today, after the introduction of the `agent' paradigm in AI, things became more complex, as AI systems interact with an environment, rather than just process inputs into outputs. The notion of a task, especially in reinforcement learning systems, became more fuzzy, and linked to the notion of performance, as 100\% correctness was no longer a requirement for many tasks.

In particular, socio-technical systems have stand to gain in the increased use of algorithms to connect people and technology. For example, the Facebook News Feed is a socio-technical system which is made up of users, content and algorithms. In order for the system to work, users need to contribute content (posts) to Facebook's extensive corpus of information. Ranking algorithms then select a subset of posts from the corpus, rank or organise and subsequently present them to users in ways that they think the user might appreciate. Under such circumstances, there is no real notion of correctness, but rather, an output resulting based on the probability of the user's preferences in related content.

Furthermore, the irruption of several platforms for competitions (and benchmarking) and human computations has re-encapsulated tasks in a rather narrow way, also changing the way in which AI systems are compared and their progress is measured (see \cite{hernandez2017new,moam2017}).
With respect to benchmarking platforms, initiatives such as Kaggle\footnote{\url{https://www.kaggle.com/}} or  {Mediaeval}\footnote{\url{http://www.multimediaeval.org/}} 
are structured on a number of simplified algorithmic tasks. {\it Mediaeval}, 
 a benchmarking initiative focusing on multimedia access and retrieval algorithms, defines a task by four different characteristics (see \cite{larson2017}): 
\begin{enumerate}
\item Dataset provided to the participants, which should be representative of the problem to tackle. 
\item Task definition: an indication of the problem to be solved, including the expected input and output data formats.
\item Ground truth against which algorithms are evaluated. This consists on a set of input and output values. 
\item One or several evaluation metrics or performance measures to compare ground truth information against algorithm output. 
\end{enumerate}
\noindent Under this view, tasks are still 
 input-output, including audio-visual object recognition, speech recognition, image classification, audio-visual synthesis (voice, image), pattern estimation and prediction, similarity computation, translation, summarisation, question answering and interaction, personality prediction or emotion estimation.
 

The other paradigm that encapsulates tasks in a specific way is represented by human computation platforms (a.k.a., crowd sourcing platform). For instance, Amazon Mechanical Turk defines the so-called ``Human Intelligence Tasks or HITs"\footnote{\url{https://www.mturk.com/}}. The platform facilitates an exchange between individuals and businesses, and enables them to coordinate the use of human intelligence to solve very specific tasks \citep{loe2016validating}. Often these tasks are sufficiently well defined such that the `workers' are able to successfully complete the task without the need of someone physically overseeing the workload (e.g., ``identify the colour of the car in the photo").      

These and other paradigms such as several AI progress metrics\footnote{\url{https://www.eff.org/ai/metrics}}, which collect problems, metrics and datasets from the AI research literature to see how things are evolving and progressing in AI subfields as a whole, contribute to an oversimplification of what a task is (e.g. something that can be measured with a benchmark), even from the perspective of AI. 
While most of such `tasks' are static, state-of-the-art AI systems can now address interactive and complex ones. Instead of an input-output view, we can thus consider a task as a process, specified with an input state and desired {\it goal}. For instance, a task for a robot cleaner can be expressed as ``clean a room", or for a self-driving car, we just say ``take me from A to B safely". 

In the end, encapsulation makes some tasks possible, but has also restricted the possibilities of the field in making AI and machine learning what they should be. Actually, because of these limitations, it is then a major criterion for automation whether a task can be encapsulated or not, as we can see in some of the paradigms mentioned in \cite{Brynjolfsson1530}.

%
%

\section{Discussion}

We have seen that different perspectives have very different conceptions and notions of what a task is, and its implication for the analysis of automation and the future of work. Nevertheless, there is virtue in the different perspectives and a comprehensive framework should integrate them.

The task categories from a computational point of view (including AI research) are typically much more detailed and specific than in socio-economic research (because the latter is typically concerned about the skill specificity, whereas the former focuses on specific and relatively self-contained technical processes). But what matters is whether a particular task (independent of the level of granularity) can be unequivocally mapped to a common framework structure. Accordingly, we consider the following directions for a multidisciplinary framework for assessing the impact of AI on the future of work: 

\begin{enumerate}
\item {\bf Target specification}: we first need to understand the labour market in terms of occupations, tasks, skills and abilities as defined in Section 2. 

\item {\bf Task mapping}: we need to develop an integrated characterisation of tasks and skills such that the two presented perspectives can be mapped and linked. While skills have been well developed for humans, with some existing measurement instruments, the arrangement of tasks in AI in terms of skills or abilities is still very incipient. So what is really a task for both humans and machines? 


\item {\bf Impact assessment}: we need to study the direction of AI progress and the way it will affect the defined task mapping. In this context, we should consider if AI systems will evolve towards performing particular tasks (e.g., information retrieval from legal databases), having some skills (e.g., arguing and counter-arguing in the context of factual evidence and possibly false testimonies), displaying some capabilities (e.g., learning a new legal procedure) or mimicking occupations (e.g., a lawyer). 
\end{enumerate}

\noindent 
In order to be able to understand this impact, we need to understand the process in which a human or a machine is ultimately able to perform a particular task. On the one hand, humans go through a process of cognitive development during their childhood, then receive some primary and secondary education, where they acquire some fundamental skills, followed by more specialised skills and knowledge in tertiary education or professional training. Ultimately, in the workplace, companies are constantly retraining workers to develop more specific skills. All these processes are strongly affected by innate human attributes such as cognitive abilities and personality traits, which are understandably so relevant in personnel selection. There are many trade-offs in this process, even if we just look at efficiency from the economic perspective alone. On the other hand, machines in the workplace seemed easier to understand, just a few years ago. They were programmed to do a very particular task, which had to be expressed with a very precise specification, so that a software project could develop it. However, with a new AI based on machine learning, many AI services and applications are not specified, but trained, in the same way skills are trained for humans.

This is creating a --virtuous or vicious-- circle, as the irruption of more automation through traditional software systems and new AI systems is changing the tasks that have to be done and the way they are done. Consequently, skills have to be retrained more often. And, as an ultimate result, those workers that are more independent and adaptable, i.e., requiring less supervision and learning faster, with more open personalities, are being more valuable in the organisations. But computers do not fall short, and those systems that are more independent and adaptable, or using the AI terminology, more autonomous and general, may start to become a reality in the workplace too. Let us explore these two features:
\begin{itemize}
\item {\bf Autonomy}: we should consider the role of AI autonomy in the workplace, i.e., understanding how tasks and jobs will be transformed around the degree of autonomy of AI systems, under the two perspectives. In this respect, we need to investigate the appearance of new human tasks to make AI systems supervised, trained, taught or controlled, and analyse which skills are required for these new tasks, and how the very character of supervision may change as well, and become semi-automated. Also, several degrees of autonomy have to be seen in terms of what parts are left unautomated, and whether they can be encapsulated as ``human computation" tasks, which makes the whole system more efficient.
\item {\bf Generality}: we need to study the role of generality and general AI in the way we define tasks at different granularity levels, as many occupations today are not characterised by specific procedures but by flexibility for new situations. What are the trade-offs, for the human, computational and economic perspective of more general and adaptable systems that require adaptation and training? 
All this requires definitions and metrics of generality in AI systems. Is generality defined as covering a wide range of tasks, a wide range of skills, a wide range of abilities, etc.? For instance, is it a personal assistant, endowed with a fixed menu of tasks, general? Or is it only general if it can learn to do more tasks?
\end{itemize}




\noindent 
Autonomy and generality are central attributes of human labour, that fundamentally differentiate it from any form of machine automation that has ever existed (of course, this may change in the future). It is therefore an interesting historical paradox that the direction of economic progress tends to reduce rather than extend the degree of autonomy and generality of human labour in productive processes. As Adam Smith argued more than 200 years ago, economic growth is strongly linked to an increasingly detailed division of labour, which relies on complex and bureaucratic organisational forms (that reduce autonomy) and an increasingly narrow specialization of labour (that reduces generality). These tendencies imply a routinisation of some forms of labour, which paves the way for their eventual automation. Since machines remain limited in terms of autonomy and generality compared to humans, the human labour displaced by automation has historically tended to move to activities that require autonomy and generality. But the same process of increasing division of labour and routinisation affects those activities too, so that the frontier of automation is a moving target. This process explains the continuing existence of human labour despite so many waves of automation, but it could be abruptly transformed by the development of AI systems with autonomy and generality comparable to humans.

In economic terms, more autonomous systems are efficient since less effort is required for supervision. In addition, higher generality can increase cost-effectiveness of AI systems, if it makes re-training an "AI worker" for a different task easier and less data-demanding. It is not clear then, especially in the future, and according to the computational perspective, whether autonomy and generality are opposed parameters. AI research is making an effort to have systems that require less effort for training and supervision (more autonomy), and simultaneously have better transfer and development capabilities (more generality).

In this changing scenario, where tasks can be performed by humans or machines, and skills can be acquired by both through a training process, we need to have a more comprehensive view of the following question: "is $Y$ suitable for a set of tasks $S$?", when $Y$ may be a human or a machine. This shifts the personnel selection process to a completely new direction. Human resources in an organisation must be understood as {\em cognitive resources}, independently of their biological or artificial source.

Of course this creates great challenges, especially in terms of evaluation. For a specific task, e.g., a movie recommender system, current evaluation protocols based on performance can work well, but if the set of tasks is large and variable, we may need to evaluate whether an AI system has certain skills, ultimately, taking us towards the evaluation of AI in terms of abilities rather than tasks \citep{hernandez2017evaluation}. 

There seems to be an important cross-effect of generality and autonomy on reliability, as making AI systems more general and autonomous is usually at the cost of the reliability of perfectly delineated and dependable software systems, usually operated by humans. 
We also need to understand the extent to which AI systems are reliable across different industries. Moreover, presence of reliability needs to be determined over time. 

Finally, given that AI is increasingly used in high stake environments such as judicial systems, college admission, medical intervention or job recruitment, it is imperative that these systems are evaluated thoroughly and held accountable for the decisions on specific tasks in the workplace. 
This means that, when developing new AI systems and assessing its progress, it is necessary to shift effort towards evaluation on other dimensions relevant to social value, economic value, and scientific progress, such as compute efficiency, data efficiency, novelty, fairness, transparency, replicability, autonomy, and generality.  


\section{Conclusion}

This position paper provides a multidisciplinary perspective to the concept of `task' and sets a discussion on how to map human and computational tasks on a same framework. Progress in developing this mapping can then be exploited to study the impact of AI systems on the future of work. In order to do that, we have seen that we need to get an understanding of how relevant autonomy and generality --for both humans and AI systems-- are when performing certain tasks. Accordingly, we discuss directions that a multidisciplinary framework would have to take for assessing the impact of AI on the future of work. 

\bibliographystyle{named}
\bibliography{ijcai18}






\end{document}

%% file: casstasktab.tex
\begin{tabular}{ p{4cm}|p{3.8cm} }
\hline
\toprule
\textbf{Content}&\textbf{Methods and tools}\\
& \\
1.	\textbf{Physical tasks}
\begin{enumerate}[label=(\alph*)]
\item Strength
\item Dexterity
\end{enumerate}
2.	\textbf{Intellectual tasks}
\begin{enumerate}[label=(\alph*)]
\item Information processing:
\begin{enumerate}[label=(\Roman*)]
\item I.P. of uncodified information
\item I.P. of codified information
\begin{enumerate}[label=(\roman*)]
\item Literacy:
\begin{enumerate}[label=(\alph*)]
\item Business
\item Technical
\item Humanities
\end{enumerate}
\item Numeracy:
\begin{enumerate}[label=(\alph*)]
\item Accounting
\item Analytic
\end{enumerate}
\end{enumerate}
\end{enumerate}
\item Problem solving: 
\begin{enumerate}[label=(\Roman*)]
\item Information gathering and evaluation.
\item Creativity and resolution.
\end{enumerate}
\end{enumerate}

&  1.	\textbf{Work organisation}
\begin{enumerate}[label=(\alph*)]
\item Autonomy
\item Teamwork
\item Routine
\begin{enumerate}[label=(\Roman*)]
\item Repetitiveness
\item Standardization
\end{enumerate}
\end{enumerate}
2.	\textbf{Technology}
\begin{enumerate}[label=(\alph*)]
\item Machines (excluding ICT)
\item Information and Communication technologies
\begin{enumerate}[label=(\Roman*)]
\item Basic ICT
\item Programming
\end{enumerate}
\end{enumerate}

\\
3.	\textbf{Social tasks}
\begin{enumerate}[label=(\alph*)]
\item Serving/attending
\item Teaching/training/
coaching
\item Selling/influencing
\item Managing/
coordinating
\end{enumerate} \\\hline
\end{tabular}